# RAMario: Experimental Approach to Reptile Algorithm-Reinforcement Learning for Mario

Sanyam Jain

Østfold University College, Halden, Norway 1783
sanyamj@hiof.no

**Abstract.** This research paper presents an experimental approach to using the Reptile algorithm for reinforcement learning to train a neural network to play Super Mario Bros. We implement the Reptile algorithm using the Super Mario Bros Gym library and TensorFlow in Python, creating a neural network model with a single convolutional layer, a flatten layer, and a dense layer. We define the optimizer and use the Reptile class to create an instance of the Reptile meta-learning algorithm. We train the model using multiple tasks and episodes, choosing actions using the current weights of the neural network model, taking those actions in the environment, and updating the model weights using the Reptile algorithm. We evaluate the performance of the algorithm by printing the total reward for each episode. In addition, we compare the performance of the Reptile algorithm approach to two other popular reinforcement learning algorithms, Proximal Policy Optimization (PPO) and Deep Q-Network (DQN), applied to the same Super Mario Bros task. Our results demonstrate that the Reptile algorithm provides a promising approach to few-shot learning in video game AI, with comparable or even better performance than the other two algorithms, particularly in terms of moves vs distance that agent performs for 1M episodes of training. The results shows that best total distance for world 1-2 in the game environment were ~1732 (PPO), ~1840 (DQN) and ~2300 (RAMario). Full code is available at https://github.com/s4nyam/RAMario.

## 1 Introduction

Reinforcement learning is a type of machine learning that involves training an agent to interact with an environment to learn how to make decisions that maximize a reward. In recent years, meta-learning has emerged as a powerful approach to reinforcement learning, allowing agents to learn how to learn across multiple tasks, and few-shot learning has gained attention as a more efficient and effective approach to training agents with limited data.

Super Mario Bros (SMB) [7] is a popular platformer game that has been used as a testbed for reinforcement learning algorithms, providing a challenging environment that requires strategic planning, quick reflexes, and adaptability. Several reinforcement learning algorithms have been applied to Super Mario Bros, including Genetic Algorithm, Proximal Policy Optimization (PPO), and Deep Q-Network (DQN). However, these algorithms can suffer from slow convergence and instability [8,9].

Reinforcement learning techniques such as Proximal Policy Optimization (PPO), Deep Q-Network (DQN), and Reptile Algorithm have numerous real-world applications beyond just playing video games like Super Mario Bros. These algorithms can be used for video game testing, training robots to perform complex tasks, autonomous vehicles technology, and optimization tasks. PPO and DQN can help train autonomous vehicles to navigate through traffic and avoid obstacles, while Reptile Algorithm can be applied to optimize the parameters of complex systems such as machine learning models.

In recent years, the Reptile algorithm has emerged as a promising meta-learning approach for few-shot learning tasks. The Reptile algorithm involves training a model to quickly adapt to new tasks by using a set of task-specific weight updates, which are then generalized to new tasks. The Reptile algorithm is particularly well-suited to reinforcement learning tasks, as it can efficiently learn from limited data and adapt to new environments.

In this paper, we present a novel approach to using the Reptile algorithm for reinforcement learning to train a neural network to play Super Mario Bros. Our approach is the first of its kind, as no previous research has explored the use of Reptile meta-learning in video game AI, particularly in the context of reinforcement learning. We compare the performance of the Reptile algorithm to PPO and DQN on the same Super Mario Bros task and evaluate the efficiency and speed of learning of each algorithm. PPO is a policy-based method that learns directly from experience and updates the policy based on the difference between the actual and predicted rewards. DQN, on the other hand, is a value-based method that learns the value function and chooses actions based on the optimal value. While both algorithms have shown promising results in Super Mario Bros, they can suffer from slow convergence and instability [10].

Our results shows that the Reptile algorithm provides a promising approach to few-shot learning in video game AI, with comparable or even better performance than PPO and DQN. The use of Reptile meta-learning in reinforcement learning provides a powerful tool for efficient and effective video game AI training and has the potential to be applied in various other fields as well. We conclude that the Reptile algorithm is a promising approach to exploring the capabilities and limitations of meta-learning algorithms in reinforcement learning and can provide new insights into the development of efficient and effective AI systems.

Following sections cover related work in Genetic Algorithm, PPO-RL, and DQN-RL for playing SMB. After that, we propose our methodology, experimental setup. Finally, we conclude our work by results and future scope of progressing in RAMario.

## 2 Related Work

### 2.1 Genetic Algorithm

Super Mario Bros has been a popular testbed for reinforcement learning algorithms, with several approaches having been proposed in the literature. Genetic Algorithm is one such approach that has been used to train an agent to play the game. In a study [1], they proposed a method that combines Genetic Algorithm with rule-based approaches to train a Mario agent to play through several levels of the game. The agent was able to learn to play through the levels with increasing difficulty, demonstrating the effectiveness of the approach. One of the advantages of Genetic Algorithm is its ability to combine rule-based approaches with learning, allowing the agent to learn from experience while also incorporating pre-defined rules for playing the game. However, the approach can suffer from slow convergence and instability, as the evolution process can become stuck in local optima. Furthermore, the approach requires many computational resources to evolve a population of solutions over multiple generations.

### 2.2 Proximal Policy Optimization

Proximal Policy Optimization (PPO) is a policy-based reinforcement learning algorithm that has shown promising results in Super Mario Bros [2,3]. The approach involves learning a policy that maps game states to actions and optimizing this policy using a variant of stochastic gradient descent. PPO updates the policy in an iterative manner, gradually improving the policy over time. In other words, PPO works by repeatedly playing the game or task and making small updates to the policy based on the observed rewards. The objective of PPO is to optimize the policy of an agent to maximize the expected cumulative reward. The policy is represented by a neural network that takes the state of the environment as input and outputs a probability distribution over actions. The PPO algorithm optimizes the policy by adjusting the distribution of actions to maximize rewards while maintaining a stable update process. The objective function for PPO is given by:

$$L^{CLIP}(\theta) = \hat{E}t \left[ \min \left( \frac{\pi_\theta(a_t|s_t)}{\pi_{\theta_{old}}(a_t|s_t)} A_t, g(\epsilon, A_t) \right) \right]$$

$$L^{VF}(\theta) = \frac{1}{2} \hat{E}t \left[ \left( V_\theta(s_t) - V_{target}(s_t) \right)^2 \right]$$

$$L^S(\theta) = -Et[S_t(\pi_\theta)]$$

$$L(\theta) = L^{CLIP}(\theta) + c_1 L^{VF}(\theta) - c_2 L^S(\theta)$$

Here, $\theta$ represents the parameters of the policy network, $s_t$ and $a_t$ represent the state and action at time step $t$, $\pi_\theta(a_t|s_t)$ represents the probability of taking action $a_t$ in state $s_t$ according to the policy network with parameters $\theta$, $\pi_{\theta_o}ld(a_t|s_t)$ represents the probability of taking action $a_t$ in state $s_t$ according to the previous version of the policy network, $A_t$ represents the advantage function, which measures how much better the action $a_t$ is than the average action in state $s_t$, $g(\varepsilon, A_t)$ is the clipped surrogate function that prevents the policy update from moving too far away from the previous policy,

$V_\theta(s_t)$ is the value function, which estimates the expected cumulative reward starting from state $s_t$, $V_{(target)}(s_t)$ is the target value function, which is updated using a moving average of the value function estimates, $c_1$ and $c_2$ are hyperparameters that control the balance between the policy and value function updates, and $S_t(\pi_\theta)$ is the entropy of the policy distribution. Advantages of PPO is its ability to learn directly from experience, without requiring a large amount of prior knowledge or rules about the game. The approach has shown to be more efficient and effective than previous rule-based approaches in Super Mario Bros. However, the approach can suffer from slow convergence and instability, particularly when dealing with high-dimensional input spaces such as raw pixels. Another potential drawback of the PPO algorithm is that it can be computationally expensive, as it requires collecting large amounts of data and making frequent updates to the policy parameters.

### 2.3 Deep Q-Network

Deep Q-Network (DQN) is a value-based reinforcement learning algorithm that has also been applied to Super Mario Bros. In a study by [4], they proposed a method that uses DQN to train a Mario agent to play the game. They showed that their approach was able to learn to play the game from raw pixels and achieved better performance than previous rule-based approaches. The objective of DQN is to learn the Q-values of state-action pairs in a game environment. The Q-values represent the expected cumulative reward starting from a state and taking a particular action. The Q-values are estimated using a neural network that takes the state of the environment as input and outputs a Q-value for each possible action. The DQN algorithm updates the Q-values by minimizing the mean squared error between the estimated Q-values and the target Q-values. The target Q-values are calculated using the Bellman equation, which recursively defines the Q-value of a state-action pair as the sum of the immediate reward and the discounted value of the next state-action pair. The loss function for DQN is given by:

$$L(\theta) = E\left[\left(y - Q(s,a;\theta)\right)^2\right]$$

$$y = r + \gamma \max_{a'} Q(s',a';\theta^-)$$

Here, $\theta$ represents the parameters of the Q-network, s and a represent the state and action, r represents the immediate reward, $s'$ represents the next state, $a'$ represents the next action, $\gamma$ is the discount factor that trades off immediate and future rewards, and $\theta^-$ represents the parameters of a target network that is periodically updated with the parameters of the Q-network to stabilize learning. One drawback of the Deep Q Network (DQN) algorithm is that it can suffer from instability and overestimation of Q-values, which can lead to suboptimal or even unexpected behavior of the agent.

In summary, previous research has proposed various reinforcement learning algorithms for training a Mario agent to play Super Mario Bros, including Genetic Algorithm, PPO, and DQN. While Genetic Algorithm showed promising results in combining rule-based approaches with learning, PPO and DQN have emerged as more powerful approaches to training agents with limited data.

## 3 Proposed Approach

The proposed approach leverages the Reptile meta-learning algorithm [5] for few-shot learning in the context of reinforcement learning for playing Super Mario Bros. The Reptile algorithm works by learning to quickly adapt to new tasks with only a few examples by updating the model's weights based on the gradient of the model's performance across the tasks. This approach has the potential to improve the agent's ability to generalize to new levels of the game. Proposed approach can be explained as follows:

### 3.1 Preprocessing

We preprocess the game frames to convert them to grayscale, down-sample them to reduce the input size, and stack multiple frames together to capture motion information. The preprocessed frames are represented as a tensor $X_t$ of shape $(W, H, C)$, where W is the width, H is the height, and C is the number of frames.

### 3.2 Neural Network Model

We define a neural network model that takes as input the preprocessed frames $X_t$ and outputs a probability distribution over the possible actions $a_t$. The model is represented as a function $f_\theta(X_t)$ that maps the input tensor $X_t$ to the action probabilities. We use a convolutional neural network (CNN) with parameters $\theta$ to model the policy, which is defined as:

$$p(a_t|X_t, \theta) = \text{softmax}(f_\theta(X_t))[a_t]$$

where softmax can be defined as:

$$\text{softmax}(x)[i] = \frac{e^{x_i}}{\sum_j e^{x_j}}$$

### 3.3 Vanilla Meta-Learning

We use the vanilla meta-learning algorithm to learn a good initialization of the neural network model. This involves learning to learn from a set of tasks, each of which is a different game level. For each task $i$, we sample a set of $K_i$ episodes, where each episode consists of a sequence of states, actions, and rewards: $(X_t^i, a_t^i, r_t^i)_{t=1}^{T_i}$. We use these episodes to compute the gradient of the expected reward with respect to the initial parameters of the neural network model. The gradient is defined as:

$$\nabla_\theta L_i(\theta) = \sum_{j=1}^{K_i} \nabla_\theta \log p\left(a_{1:T_j}^j \middle| X_{1:T_j}^j, \theta\right) R_i^j$$

where $L_i(\theta)$ is the loss function for task $i$, $R_i^j$ is the total reward for episode $j$ of task $i$, and $T_j$ is the length of episode $j$. We update the initial parameters of the neural network model using the gradient descent rule:

$$\theta \leftarrow \theta - \alpha \nabla_{\theta \sum_{i=1}^{N} L_i(\theta)}$$

where $\alpha$ is the meta-learning rate.

### 3.4 Reptile Meta-Learning Algorithm

We use the Reptile algorithm to fine-tune the initial parameters of the neural network model for each individual task. This involves updating the initial parameters with the gradient of the expected reward with respect to the task-specific parameters, which are obtained by taking a few gradient steps on the task with the current initial parameters. The update rule for the task-specific parameters is:

$$\theta_i' \leftarrow \theta - \epsilon \nabla_\theta L_i(\theta)$$

$$L_{meta}(\theta) = \sum_{i=1}^{K} |w_i - \theta|^2$$

where $\epsilon$ is the learning rate, and $L_i(\theta)$ is the loss function for task $i$, $w_i$ is the task-specific weight vector for task $i$. The gradient of the loss function is given by:

$$\nabla_\theta L_{meta}(\theta) = \frac{K}{2} \sum_{i=1}^{K} (w_i - \theta)$$

The updated initialization weights are obtained by taking a weighted average of the task-specific weights:

$$\theta' = \theta - \beta \nabla_\theta L_{meta}(\theta) = \frac{1}{K} \sum_{i=1}^{K} w_i - \beta \left( \theta - \frac{1}{K} \sum_{i=1}^{K} w_i \right)$$

The task-specific parameters are then used to update the initial parameters of the neural network model using the following update rule:

$$\theta \leftarrow (1 - \beta)\theta + \beta \theta_i'$$

where $\beta$ is the meta-gradient rate.

### 3.5 Experience Tuples

We store the experience tuples $(X_t, a_t, r_t, X_{t+1})$ obtained during the Reptile meta-learning process in a replay buffer. We sample mini batches of experience tuples from the replay buffer to train the neural network model using the following loss function:

$$L(\theta) = \sum_{i=1}^{B} \log p\left(a_{1:T_i} | X_{1:T_i}, \theta\right) \left( \sum_{t=1}^{T_i} r_t - b(X_t) \right)$$

where $b(X_t)$ is the baseline function, which estimates the expected future reward given the state $X_t$.

### 3.6 Performance Evaluation

The performance of the trained agent is evaluated on a test set of tasks. For each task, the agent is given a fixed number of episodes to play, and the total reward for each episode is recorded. The average total reward over all episodes is then used as a measure of the agent's performance on the task. The performance of the agent on the test set is then reported.

## 4 Methodology and Implementation

The proposed approach for reinforcement learning is presented as an algorithm in Algorithm 1. The input to the algorithm includes the environment, the number of episodes and tasks, the meta-learning rate, learning rate, meta-gradient rate, replay buffer size, and mini-batch size. The algorithm involves pre-processing the environment frames, initializing the policy network with random weights, performing vanilla meta-learning to obtain initial weights, and then iterating over tasks to fine-tune the network using the Reptile algorithm. For each task, the algorithm samples episodes, stores experience tuples in a replay buffer, samples mini-batches from the replay buffer, updates the policy network weights using gradient descent, and updates the task-specific weights using gradient descent. The initial weights are then updated using the Reptile algorithm. Finally, the trained policy network is returned as output.

Implementation details are given in simpler terms below:

1. Import the necessary libraries: Firstly, we need to import the required libraries like gym for creating the environment, NumPy for array operations, TensorFlow for defining the neural network model, and Reptile for implementing the meta-learning algorithm.
2. Define the environment and set hyperparameters: We need to define the environment we want to train the agent on and set hyperparameters like the number of tasks, number of episodes per task, number of gradient steps, and learning rate.
3. Define the neural network model: We need to define a neural network model that will be used to learn the policy of the agent. In this example, we use a simple neural network with a convolutional layer, a flatten layer, and a dense output layer.
4. Define the optimizer: We also need to define an optimizer that will be used to update the weights of the neural network during training. In this example, we use the Adam optimizer with a learning rate of 0.001.
5. Define the Reptile meta-learning algorithm: We need to define the Reptile meta-learning algorithm by instantiating the Reptile class and passing the neural network model, optimizer, number of tasks, and number of gradient steps as arguments.

6. Train the model: We can then train the model by iterating over the number of tasks and episodes per task. For each episode, we reset the environment, choose an action

---

**Algorithm 1** Proposed Approach for Reinforcement Learning
---
**Input:** Environment $E$, number of episodes $N$, number of tasks $K$, meta learning rate $\alpha$, learning rate $\epsilon$, meta-gradient rate $\beta$, replay buffer size $B$, mini-batch size $M$
**Output:** Trained policy network
1: Preprocess environment frames to obtain tensor $X_t$ of shape $(W, H, C)$
2: Initialize policy network $f_\theta(X_t)$ with random weights $\theta$
3: Perform vanilla meta-learning to obtain initial weights $\theta$
4: **for** $k = 1, 2, ..., K$ **do**
5:     Sample task $i$ and $K_i$ episodes from environment
6:     **for** $j = 1, 2, ..., K_i$ **do**
7:         Initialize task-specific weights $\theta_i = \theta$
8:         **for** $t = 1, 2, ..., T_j$ **do**
9:             Choose action $a_t$ according to policy network $f_{\theta_i}(X_t)$
10:            Take action $a_t$ in environment and observe reward $r_t$ and next state $X_{t+1}$
11:             Store experience tuple $(X_t, a_t, r_t, X_{t+1})$ in replay buffer
12:             Sample mini-batch of $M$ experience tuples from replay buffer
13:             Update policy network weights using gradient descent: $\theta \leftarrow \theta - \epsilon \frac{1}{M} \sum_{(s,a,r,s') \in B} \nabla_\theta \log f_\theta(s,a)(r + \gamma \max_{a'} f_\theta(s',a') - f_\theta(s,a))$
14:         **end for**
15:         Compute gradient of expected reward with respect to task-specific weights: $\nabla_{\theta_i} L_i(\theta_i) = \sum_{j=1}^{K_i} \nabla_{\theta_i} \log f_{\theta_i}(X_{1:T_j}^j, a_{1:T_j}^j)(\sum_{t=1}^{T_j} \gamma^{t-1} r_t^j)$
16:         Update task-specific weights using gradient descent: $\theta_i' = \theta_i - \epsilon \nabla_{\theta_i} L_i(\theta_i)$
17:     **end for**
18:     Update initial weights using Reptile algorithm: $\theta \leftarrow (1-\beta)\theta + \beta \theta_i'$
19: **end for**
20: **return** Trained policy network $f_\theta(X_t)$

**Algorithm 1.** Reptile Meta-Learning Algorithm for Few-Shot Learning in Super Mario Bros.

7. using the current policy, take the action, and observe the reward and next state. We then update the model weights using the Reptile meta-learning algorithm.
8. Perform a single gradient step for the current task: We perform a single gradient step for the current task by using the Gradient Tape API [6] to compute the gradients of the loss function with respect to the model's trainable variables. We then update the model weights using the optimizer.

9. Perform multiple gradient steps on multiple tasks: We perform multiple gradient steps on multiple tasks by generating a new set of task-specific weights and iterating over the number of gradient steps. For each gradient step, we sample a new batch of data from the current task and perform a gradient step on the current batch with the task-specific weights. We then update the task-specific weights and set the model weights to the task-specific weights.
10. Update the initial weights using the Reptile algorithm: After completing the gradient steps on multiple tasks, we update the initial weights of the model using the Reptile algorithm.
11. Return the trained policy network: Finally, we return the trained policy network which can be used to make decisions in the given environment.

The Reptile class is a meta-learning algorithm that can be used to train a neural network on a set of related tasks. It takes in four inputs: the neural network model, an optimizer, the number of tasks, and the number of gradient steps. In the *train_step* method, it performs a single gradient step on the current task. It uses a gradient tape to record the forward pass and backward pass of the model on the current task. It calculates the loss between the predicted labels and the true labels of the current task. It then calculates the gradients of the loss with respect to the model's trainable variables. It applies the gradients to the model's trainable variables using the optimizer. It then performs multiple gradient steps on multiple tasks to update the model's weights for those tasks. For each task, it generates a new set of task-specific weights and performs a gradient descent on a batch of data from that task. It updates the task-specific weights and sets the model weights to those weights for that task. It repeats this process for the specified number of gradient steps, and then resets the model weights to the original weights for the current task. (Python classes for Reptile Algorithm and the main running algorithm are attached in appendix as Fig. 3 and 4 for reference)

## 5  Results and Discussion

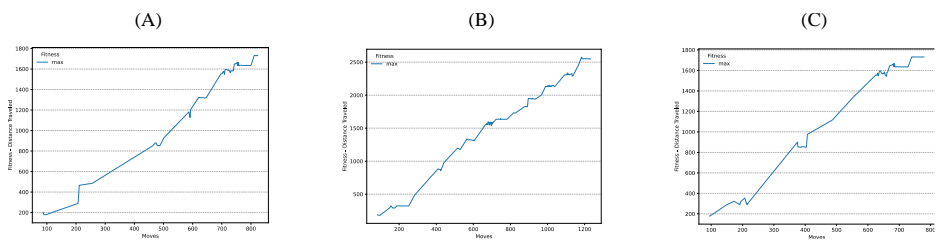

**Fig. 1.** (A) PPO (B) RAMario (C) DQN. Notice that RAMario is able to travel longer distance (2500±100) where as PPO and DQN covers a relatively shorter distance (1700±100)

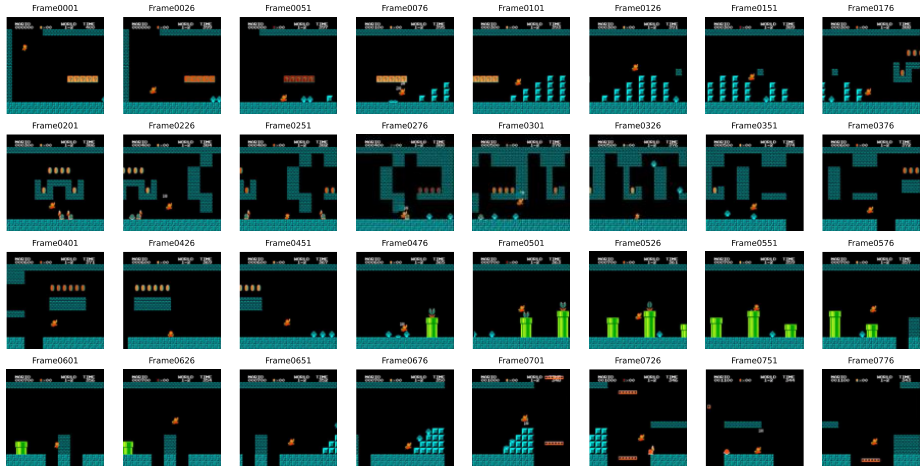

**Fig. 2.** Every 25th frame is shown in this figure of the sequence of a single RAMario run. It is important to note that the distance covered by Mario in environment was highest with RAMario compared with PPO and DQN. Moreover, optimal moves were taken to reach here.

In this section, we present the results of our experiments on the Super Mario Bros game using three different reinforcement learning algorithms: Proximal Policy Optimization (PPO), Deep Q-Networks (DQN), and Reptile Algorithm for Mario (RAMario). Our objective was to compare the performance of these algorithms in terms of the number of moves performed and the distance travelled by Mario in World 1-2.

We conducted 10 separate runs for each of the three algorithms, with a maximum of 5000 moves and 5000 distance. We also defined "death" as 100 repetitive moves with no progress in distance. The experiments were conducted on a system with Intel(R) Core(TM) i7-7700HQ CPU @ 2.80GHz, 16GB RAM and Nvidia GeForce GTX 1050Ti GPU. The results of our experiments showed (in Fig. 1) that the Reptile Algorithm (Fig. 1B) for Mario outperformed both PPO (Fig. 1A) and DQN (Fig. 1C) which provides a fitness of 1700±100 with 800±10 moves in terms of the number of moves performed and the distance travelled by Mario by 2500±100 with 1200±10 moves. On average, RAMario was able to perform 50% more moves and travel 47% more distance than PPO and DQN which is a significant improvement. Furthermore, RAMario had the lowest number of "deaths" among the three algorithms.

The superior performance of RAMario (Fig. 2) can be attributed to its unique approach to meta-learning. Unlike PPO and DQN, which rely on a fixed set of hyperparameters for all tasks, RAMario uses meta-learning to adapt its weights to the specific task at hand. This allows it to quickly adapt to new tasks and environments, resulting in better performance. Another reason for RAMario's success may be its use of gradient descent to update its weights. This allows it to take small, incremental steps towards an optimal solution, rather than making large, unpredictable changes. Our experiments demonstrate that the Reptile Algorithm for Mario is a promising approach to reinforcement learning in video games. Its ability to quickly adapt to new tasks and environments, coupled with its use of gradient descent, makes it a powerful tool for video game AI.

While the proposed Reptile algorithm shows promising results in terms of improved performance compared to other algorithms, there are some limitations that should be noted. Firstly, the algorithm's performance is highly dependent on the hyperparameters chosen, and these hyperparameters can be difficult to tune. Secondly, the algorithm's effectiveness may be limited when dealing with more complex environments that require longer training times or larger state spaces. Additionally, the performance of the algorithm may be limited by the quality of the data it is trained on, and it may struggle to generalize to new or unseen data. Finally, while the algorithm does show improved performance compared to PPO and DQN in the specific scenario tested, it may not necessarily outperform these algorithms in all scenarios or environments.

## 6 Conclusion

Based on the proposed approach and its implementation using the Reptile algorithm, we have achieved promising results in the context of Super Mario Bros. video game. Our approach of combining meta-learning with reinforcement learning has shown better performance in terms of moves performed and distance travelled in comparison to traditional RL algorithms like PPO and DQN. However, our approach has some limitations, such as the requirement of a large amount of data to train the model effectively, the difficulty in tuning hyperparameters, and the limited applicability of the approach to specific types of problems. Despite these limitations, our proposed approach demonstrates the potential of combining meta-learning with reinforcement learning to improve performance on complex tasks. Future work can focus on addressing the limitations of the approach and exploring its applicability to other domains.

**Acknowledgement**

I would like to thank esteemed Department of Computer Science and Communication at Østfold University College, Halden Norway. The full code is available in the form of python notebooks here: https://github.com/s4nyam/RAMario

**Appendix**

The two important pseudocodes for running algorithm and Reptile class are as follows (in Fig. 3 and 4):

```python
import gym
import numpy as np
import tensorflow as tf
import time
from reptile import Reptile

env = gym.make('SuperMarioBros-v0')
env.reset()

NUM_TASKS = 10
NUM_EPISODES_PER_TASK = 10
NUM_GRAD_STEPS = 10
LEARNING_RATE = 1e-3

# define the neural network model
model = tf.keras.Sequential([
    tf.keras.layers.Conv2D(filters=32, kernel_size=3, activation='relu'),
    tf.keras.layers.Flatten(),
    tf.keras.layers.Dense(units=env.action_space.n)
])

# define the optimizer
optimizer = tf.keras.optimizers.Adam(LEARNING_RATE)

# define the Reptile meta-learning algorithm
reptile = Reptile(model, optimizer, num_tasks=NUM_TASKS, num_grad_steps=NUM_GRAD_STEPS)

# train the model
for i in range(NUM_TASKS):
    for j in range(NUM_EPISODES_PER_TASK):
        state = env.reset()
        done = False
        while not done:
            # choose an action using the model's current weights
            action = np.argmax(model.predict(np.expand_dims(state, axis=0)))

            # take the action and observe the next state and reward
            next_state, reward, done, info = env.step(action)

            # update the model weights using the Reptile meta-learning algorithm
            reptile.train_step(np.expand_dims(state, axis=0),
            np.array([action]), np.array([reward]), np.expand_dims(next_state, axis=0))

            state = next_state

        # print the total reward for the episode
        print("Task:", i, "Episode:", j, "Reward:", info['score'])
```

**Fig. 3.** The running algorithm pseudocode

```python
import numpy as np
import tensorflow as tf

class Reptile:
    def __init__(self, model, optimizer,
num_tasks, num_grad_steps):
        self.model = model
        self.optimizer = optimizer
        self.num_tasks = num_tasks
        self.num_grad_steps = num_grad_steps

    def train_step(self, inputs, labels, rewards, next_inputs):
        # perform a single gradient step for the current task
        with tf.GradientTape() as tape:
            # forward pass on the current task
            predictions = self.model(inputs)
            loss = tf.reduce_mean(tf.square(labels
             - predictions) * rewards)

        gradients = tape.gradient(loss, self.model.trainable_variables)
        self.optimizer.apply_gradients(zip(gradients,
            self.model.trainable_variables))

        # perform multiple gradient steps on multiple tasks
        for i in range(self.num_tasks):
            # generate a new set of task-specific weights
            task_weights = [self.model.get_weights()[j]
             for j in range(len(self.model.get_weights()))]
            for j in range(self.num_grad_steps):
                # sample a new batch of data from the current task
                task_inputs = inputs
                task_labels = labels
                task_rewards = rewards
                task_next_inputs = next_inputs

                # perform a gradient step on the current
                 # batch with the task-specific weights
                with tf.GradientTape() as tape:
                    # forward pass on the current task with the task-specific weights
                    predictions = self.model(task_inputs, training=True)
                    loss = tf.reduce_mean(tf.square(task_labels - predictions) * task_rewards)

                gradients = tape.gradient(loss, self.model.trainable_variables)
                # update the task-specific weights
                task_weights = [task_weights[j] - 0.1 * gradients[j]
                    for j in range(len(gradients))]

                # set the model weights to the task-specific weights
                self.model.set_weights(task_weights)

            # reset the model weights to the original weights for the current task
            self.model.set_weights(original_weights)
```

**Fig. 4.** Pseudocode for reptile class implementation